# Interpretation of Chest x-rays affected by bullets using deep transfer learning

**Shaheer Khan**
(MNS-UAM, Multan, Punjab, shaheerkhan36302@gmail.com)

**Azib Farooq**
(UET, Taxila, Punjab, azibfarooq10@gmail.com)

**Israr Khan**
(MNS-UAM, Multan, Punjab, israr.hussain@mnsuam.edu.pk)

**Muhammad Gulraiz Khan**
(UoL, Lahore, Punjab, khangulraiz5@gmail.com)

**Abdul Razzaq**
(MNS-UAM, Multan, Punjab, abdul.razzaq@mnsuam.edu.pk)

**Abstract:** The potential of deep learning, especially in medical imaging, initiated astonishing results and improved the methodologies after every passing day. Deep learning in radiology provides the opportunity to classify, detect and segment different diseases automatically. In the proposed study, we worked on a non-trivial aspect of medical imaging where we classified and localized the X-Rays affected by bullets. We tested Images on different classification and localization models to get considerable accuracy. The replicated data set used in the study was replicated on different images of chest X-Rays. The proposed model worked not only on chest radiographs but other body organs X-rays like leg, abdomen, head, even the training dataset based on chest radiographs. Custom models have been used for classification and localization purposes after tuning parameters. Finally, the results of our findings manifested using different frameworks. This might assist the research enlightening towards this field. To the best of our knowledge, this is the first study on the detection and classification of radiographs affected by bullets using deep learning.

**Keywords:** Localization, GSW, X-rays, CNN, SSD, Faster RCNN, Deep learning.

## 1. Introduction

Gunshot wounds (GSW) results in the damage of bones and, infection of wounds and other organs of a human body (MedlinePlus 2018). There are different parameters to measure the deterioration of body organs such as velocity, size, the shape of the bullet, distance covered, and the sensitivity of tissues where the bullet impacted. The size of the wound due to implanted bullet may less, greater, or equal to the size of a shell. The process of magnification while taking radiographs, the size of bullet may disturb. The estimation of bullet caliber by using x-ray can be determined in two ways, one can be to examine two radiographs at the degrees of 90. In the second approach, a micrometer device is used for comparison between the shadow of the bullet in the X-Ray and the relevant bullets having similar caliber. The former case has been done rarely in clinical practice. In a nutshell, it is difficult to measure the caliber of the bullet accurately, the mostly process depends on estimation (Messmer 1998, Prahlow 2010, Heard 2011).

Other techniques used to visualize the body parts of a human that cause hazards in the gunshot injuries. like Magnetic resonance imaging. That is why the proposed study focused on X-ray technology. Many people died daily just because of gunshot wounds, in the year 2016, deaths were 161,000 due to assault (McLean, et al. 2019). In 2017 deaths due to gunshot wounds were 39,773 only in the USA (McLean, et al. 2019). Mexico, Venezuela, Columbia, USA, Brazil, and Guatemala cover the most significant number of deaths with the invasion of gunshots approximately half of the total occurred worldwide (Collaborators 1990-2016).

An experienced radiologist takes 24 hours to interpret a radiograph. In contrast, after the full training, the deep learning models take a few seconds to generate the output or manifest the result. Thereafter much research in medical imaging using deep learning models may reduce the need of the radiologists. The other issue is that there are few radiologists over millions of people (Delbeke and Segall 2011). Concerning these issues, we need to deploy techniques that should overcome problems as mentioned earlier. The deep learning models already successfully tested on many other techniques of medical imaging which will discuss in the literature review section.

In the proposed, study we highlighted an automated diagnosis of a gunshot wound on X-rays using deep learning methods. As far as we know, the proposed research is held for the first time on X-rays of gunshot wound using a deep learning model. One reason may be the absence of a dataset. The deep learning models required data set for successful execution. That is why we manually made the replicated dataset using radiographs of the National Institute of Health (NIH) (Wang, et al. 2017). We selected the chest X-rays of normal cases from different dataset directories of NIH. We used tools to increase the number of cases for the dataset from already extracted actual images of gunshot X-rays that were less in amount and could not satisfy the need for deep learning models. During replication of images, a few chest radiographs were built that was difficult for the model to train and evaluate because of the dim level of contrast. It tests, either such cases can be detected by the model, or the model feels difficulties recognizing these regions. The purpose of this difficulty is to test the model how well it performs on ambiguous cases or how much ability the model must tackle hard situations because in surgical practices there is no place for errors.

However, the model trained on GSW chest radiographs was tested on multiple other organs such as leg, neck, and abdomen through localization techniques. What would be the effect of bullets on different organs? A few of the organs among others are given below:

### 1.1.    Neck, chest, and abdomen:

The implanted bullet to the neck of humans resulted in a dangerous situation because neck wounds cause severe injuries to the nervous system and nonstop bleeding (Tisherman, et al. 2008). It is all due to a higher amount of vital anatomical presence. Difficult tackling situations include expanding hematoma, compromise of the airway, neurological deficits, wound air bubbling, uncontrol bleeding in such situations frequent surgical treatment should be given to the patient (Shiroff, et al. 2013).

Chest injuries due to gunshot cause hemothorax (Severe bleeding), pericardial tamponade (Cardiac injury), and injuries to the nervous system (Marx, Walls and Hockberger 2013). Major organs of the body including the heart and lungs can be affected. Not every gunshot injury requires surgery. Sometimes consistency of persistent bleeding and air leakage does not recover without surgery (Meredith and Hoth 2007).

Vital organs inside the abdomen include kidneys, stomach, liver, pancreas, diaphragm, etc. Similar to the chest and neck, the gunshot injuries to the abdomens also cause severe bleeding, rupture of organs, neurological deficits, and compromise on the respiratory system.

The proposed model can locate the object in other organs of the body help locate GSW in different parts of the body. Binary classification and object localization (using bounding boxes) was performed on test cases. And finally, results were obtained using different libraries or tools.

## 2. Background

In 2015, there were almost 1 million gunshot wounds in the world from interpersonal violence (Vos, et al. 2017). In the year 2016, firearms caused 251,000 deaths worldwide, more than 209,000 in 1990 (Collaborators). In the course of these deaths, the consequences of an attack were 64% (161,000), 67,500 (27%) were resulted due to suicide, and (23,000) 9% were accidents (Collaborators 1990-2016).

More deaths (Figure 1) were due to homicide by firearm, the second major portion was suicide by firearm, and the rest were from other reasons. In Figure 2 the mortality rate of countries is given, and the USA is on the list of most affected countries by firearms. Approx. 37,200 deaths were caused by firearms in the USA, and it was highest concerning other countries. The deaths due to firearms are common in men than in women. Deaths in Pakistan were 2780 in 2016 due to firearms.

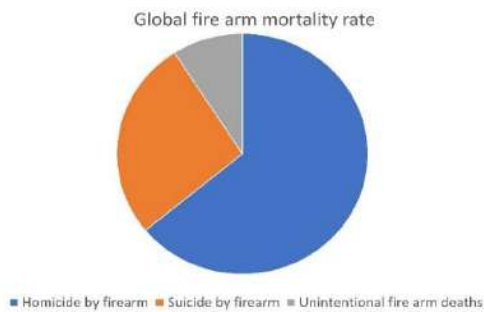

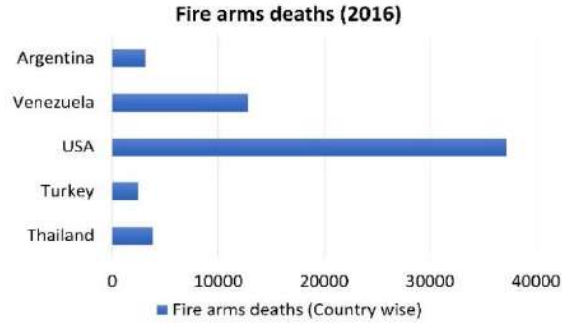

*Figure 2. Country-wise mortality rate by firearm.*

If we analyze such conditions in terms of deep learning with medical imaging, then cases of deaths can be reduced and treated in an automated manner.

In developing countries, the availability of experienced radiologists is difficult, so proper judgment is difficult to reach. So, if we use an automated way to detect bullet-affected radiographs and also locate the position of bullet from radiographs, then it can be possible to give treatment efficiently with less need of radiologist to diagnose first and then give proper treatment to the patient.

## 3. Proposed Methodology

The proposed methodology is classified into two sections: Classification and object localization. The data set of chest X-Rays are not just classified but also passes through the models of object localization. The data set that we used in the proposed study was extracted from the NIH repository of 100,000 plus chest X-Ray images (Wang, et al. 2017). This database is named "ChestX-ray8". The data set compiled more than 30,000 patients' records. For replication, few images were extracted from the NIH repository of images to make a dataset of chest X-Rays having bullet wounds in a body. Normal chest radiographs were used from the database of the National Institute of Health. So, to achieve replication, actual radiographs of bullet wounds were used to generate enough radiographs feasible to input in the model. The reason for this is to check whether the model has enough power to localize the object correctly because in the field of medical imaging there is no place for any kind of error or mistake either by humans or any computing device. Approximately 100 images were replicated to feed into the model. The 80, 20, and 10% of the dataset is reserved for training, validation, and testing. 10% of testing data based on actual radiographs of bullet wounds instead of replicated cases.

### 3.1. Classification

For the classification of images, we used a deep convolutional neural network (Krizhevsky, et al. 2012) which is considered a state-of-the-art model in the field of deep learning. Convolution Neural Networks (ConvNets) are specialized Neural Networks, which are well suited for the processing of images and had enormous success on many machine learning problems in the last few years. The type of CNN that we used was developed by the Visual geometry group of Oxford university named VGG16 (Simonyan and Zisserman 2014). Before feeding the dataset directly to the model, images are pre-processed using different augmentation techniques such as resizing, rescaling, shear range, zoom range, horizontal flip, etc. Image augmentation increases the number of images to feed in the model because the larger the dataset, the better will be the results.

Feature extraction is the process in which different patterns of images are extracted from the dataset and feedforward for further processing. For this purpose, we do not need to manually extract the features, CNN as a model does this implicitly. Filters are responsible to extract different types of features from the image dataset. Different types of filters extract different information from the images. Filters at the top layers of the CNN model catch more complex spatial information. Not all features further move forward for processing, few of them due to the threshold value of an activation function remains left for the process. Pooling is the process of downsizing the matrix information of upcoming data from the former layers. The most popular types of pooling are max pooling and average pooling. Dense layers are also known as fully connected layers in which the number of neurons are specified and activate when a feasible piece of information is received and fed. Global average pooling 2D and dropout regularization was used in dense layers. Activation after dense layers were used, such as sigmoid or softmax activation which was needed for binary or multiclass classification.

The loss was computed at the end. Loss categorizes as binary cross-entropy for binary and categorical cross-entropy for multi-class classification. Back prop needs due to the optimization of weights of the model after calculation of loss. Then during backpropagation, the weights improved in multiple iterations to make the model better than before. The proposed model used stochastic gradient descent as an optimizer with a learning rate of 0.0001 and momentum of 0.9.

The proposed study used FPT (freeze pre-train and finetune) technique. In this approach initially, we need to replace the last layer with a mini network of two fully connected Layers. Freeze all the layers that are pre-trained and train the new network on small fully connected layers, after training, weights of the network are saved. Thereafter the network finetunes again from the previous save weights on our dataset and resulted in satisfied accuracy especially on a small dataset, shown in figure 3.

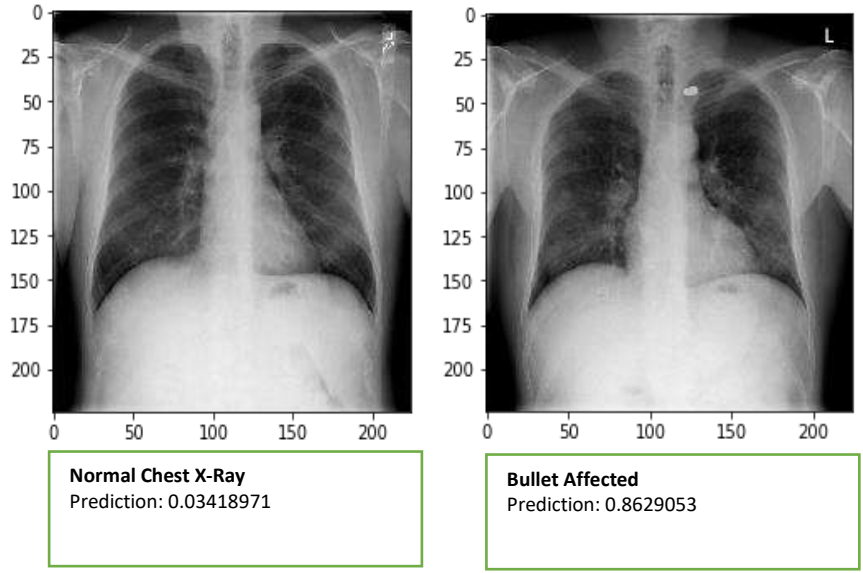

*Figure 3. Classification b/w two classes*

### 3.1.1. CAM Visualization:

Class activation heat map helps in visualizing the layer-wise outcome of a particular class after training. It manifests the focused portion by the model for a class or we can consider it as which features model learned to predict the test case for a specific class. The same approach was used in the proposed study to classify and visualize the two different classes based on notable attributes. The presence of bullets in GSW radiographs and the absence of bullets in Normal chest X-rays should be the focal point by the model. If not, then it is hard to classify because both classes are identical to each other, and bullets are the only object in images that can classify the test cases.

Figure 4 expresses the heat map of the two last layers of VGG-16 and the final superimposed image, one is of the last convolutional layer of block 5, and the other is of the pooling layer of block 5. Block 5 Conv 3 is the second last layer of VGG, but the last convolutional layer and block 5 pool is the final layer of the VGG net. Block 5 Conv 3 layer is used to obtain the CAM but if we see the block 5 pool (fig 4 (b)) it looks brighter. The reason is that in pooling the spatial dimensions are reduced and lose information. Therefore, in the block 5 pool image few lighter spots are absent that are present in block 5 Conv 3 layer i.e., in the bottom left corner of block 5 Conv 3.

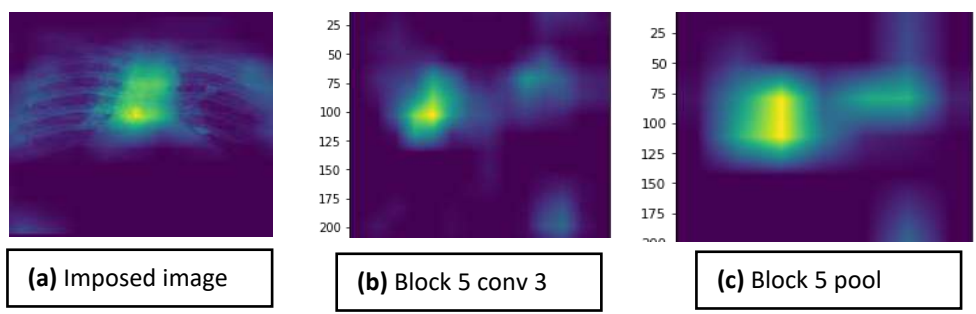

Regarding the previous statement, so what is the reason for more glint of block 5 pool layer, the reason is it resized the image to the same size of block 5 Conv 3 image (stretching) and glow the heat map due to zoom to the dimension visibility. The superimposed image in last is the placement of the actual test image on the class activation heat map. Superimposed image manifests the highlighted region of a given image and these highlighted regions are the result of training by the model on the dataset. To generate the superimposed image, the first heat map is resized according to the original image and then converted the same heat map into RGB format.

The resultant heat map is applied to the image to generate the super-imposed image as in figure 4.

## 3.2. Object Localization

Detection has progressed rapidly in recent years (Girshick et al., 2014, Ren et al., 2015, Liu et al., 2016, Redmon et al., 2016, Redmon and Farhadi 2017). The main difference between object detection and object localization is that in object localization the concern is about a single class and in object detection, multiple classes need to detect. As the proposed study focused on localization due to this, we used the term object localization.

For image classification, CNN output is a dimensional vector perhaps contains the most prominent features of all classes but in object localization, it needs to locate the position of the object too. The most prominent object features localized in it, the network can be expanded only to produce the one bounding box per image or one bounding box for each possible object category. To detect an object, we must be careful about where the object is and what is the image category. It produces a kind of chicken and eggs problem, and we need to know the shape (and category) of the thing and to learn about the location of an object, its shape is also an essential parameter.

The location and size of the bounding box are fixed, which is stored in the form of corner coordinates. The rectangle is easier to use than arbitrary polygons and many operations, such as convolution performed better on the rectangle. However, there is a sub-image in the bounding box on the rectangle and the image where the bounding box focused is classified by the model using machine learning (Girshick, et al. 2014). In object localization simple dataset of images cannot work, it needs to label the objects in the images too by making a bounding box around the object. For this, the study used labeling (Tzutalin 2020) as an annotation tool for the dataset. Labeling is an open-source free tool for annotating images graphically. It is written in Python. LabelImg assists tagging in VOC XML or YOLO format. We labeled the training and validation images in Pascal VOC XML format. In .XML (xtensible markup language) information about an image that had been labeled show in a specific format. It includes the location of an image, height, and width of an image, object label, pixel coordinates of bounding boxes, etc.

For object localization, we experimented with a dataset on different models of object detection such as Faster RCNN and SSD. Every model has its characteristics towards the detection process, such as faster RCNN has better accuracy than SSD but takes more time in execution. Faster R-CNN (Ren et al., 2015) improved after fast R-CNN after given an idea of region proposal network (RPN) that allocate fixed anchors to proposed regions from extracted feature maps use a two-dimensional CNN. Faster R-CNN is an integrated approach. The theme is to utilize convolutional layers for the generation of region proposals and detection. The authors have discovered that the feature maps that have been assigned by the model can also be explored to develop proposals for the region. The full convolutional segment of the Faster R-CNN network that gives a proposal of features is called Region proposal networks (RPN).

SSD (Liu et al., 2016) single-shot multi-box detector, takes integrated detection further. This method does not generate proposals. In addition to this, it does not involve image segments. It has only concerned with one pass of the convolutional network to generate object detections. SSD depends on feature maps and CONV filters for object detection. To restore a lack of precision, SSD has several improvements, including default boxes and multi-scale features. It handles various scales using feature maps from many different Conv layers (i.e. large and small feature maps) as entry into the classifier. Somewhat similar to the working of a sliding window, the algorithm starts with a default set of boxes surrounding. This includes the aspect ratio and scales. The calculated prediction of the boxes is based on an offset Parameter, that predicts the threshold of the bounding box from the default box. The default boundary boxes are manually checked. SSD specifies a particular value of scale for a single feature map layer. For the six predictive layers, the SSD starts with 5 different target aspect ratios: 1, 2, 3, 1/2, and 1/3. Then the default box width and height are calculated as:

$$w = scale . \sqrt{aspect\ ratio}$$
$$h = \frac{scale}{\sqrt{aspect\ ratio}}$$

Extra default box added by SSD with scale:

$$scale = \sqrt{scale . scale}$$

at next level and

$$aspect\ ratio = 1$$

These improvements allow SSD to be matched to a faster R-CNN with the use of lower resolution images, which increases speeds.

## 4. Results and Prediction

For training, we used the TensorFlow GPU library and Colab GPU (graphical processing unit) services. Initially, we cloned the repository of TensorFlow (TensorFlow 2020) and install all required packages of the COCO dataset. Thereafter download the dataset from google drive and convert all .XML files to .CSV (comma separated file) to input in TensorFlow or TF record. TF record is a binary file that helps the model to execute fast with low memory usage. It contains the images and labels of images. Afterward, we downloaded the pre-trained model from the TensorFlow object detection repository and executed three different object detection models, including SSD mobile net V1 COCO, SSD mobile net V2 COCO, and Faster RCNN Resnet101 COCO.

After complete training and saving of checkpoints, multiple unseen actual images were tested on models to check their accuracy and the level of prediction, either the bounding box draw on the correct portion or not. It must be noted that tested images not only contain images of chest radiographs but also contain other organs' radiographs like leg, head, and abdomen.

Parameters included in the evaluation are mAp (mean average precision) and average recall. Following a separate discussion of each model results incorporated in a table. The accuracy of models can be improved by performing further hyperparameter tuning.

For localization, model performance was evaluated by using Tensor-board. Tensor-board is a visualization tool used to measure different metrics of executed model and describe interactively. For the proposed study, the weights stored during the execution were used to get the model's performance from a different perspective.

### 4.1. Mean Average precision:

Mean Average Precision is just an extension of AP. You simply take the averages of all the AP scores for a certain number of queries. It is a common metric to measure the accuracy of object localization models. The more the mean average precision the better the model performance is expected.

$$mAP = \frac{1}{N} \sum_{k=1}^{N} (AP_K)$$

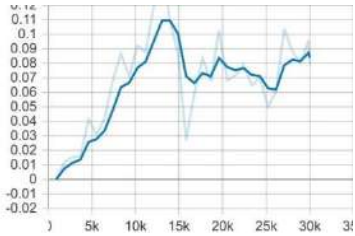 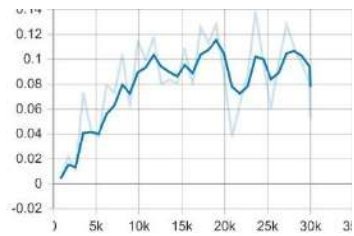 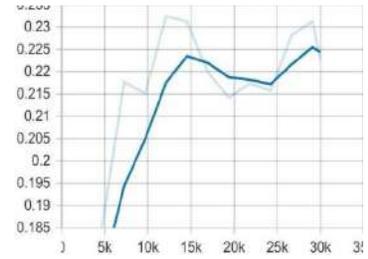

*(a) SSD Mobile Net V1 Coco*     *(b) SSD Mobile Net V2 Coco*     *(c) Faster RCNN Resnet101 Coco*

*Figure 5. mAP of Models*

### 4.2. Recall

Recall considered as the true positive (TP) rate, also termed as sensitivity, calculate the probability of ground truth objects being detected correctly.

$$Recall = \frac{True\ object\ detection}{all\ ground\ truth\ boxes} = \frac{TP}{TP + FN}$$

$$\therefore TP = True\ positive, \qquad FN = False\ Negative$$

SSD Mobilenet v1 COCO                SSD Mobilenet V2 COCO                Faster RCNN Resnet101 COCO

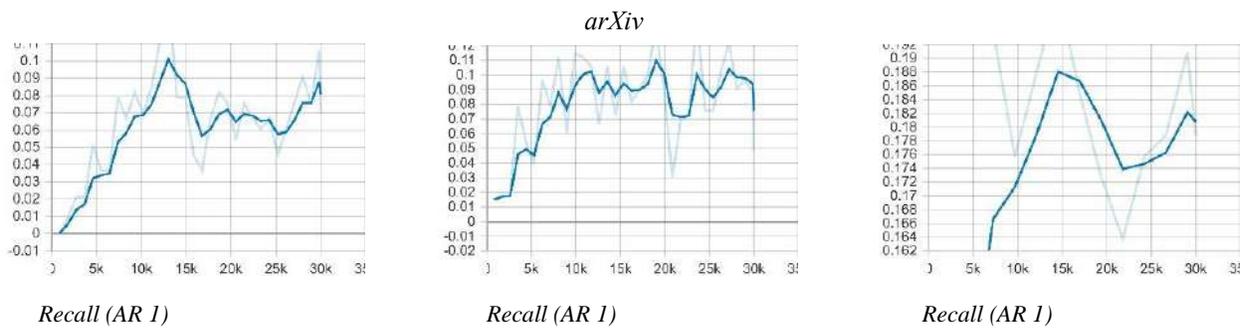

*Recall (AR 1)*     *Recall (AR 1)*     *Recall (AR 1)*

*Figure 6. AR of Models*

∴ Average Recall 1, computes the mean average recall of all images with at-most 1 detection (i.e., 0 or 1)

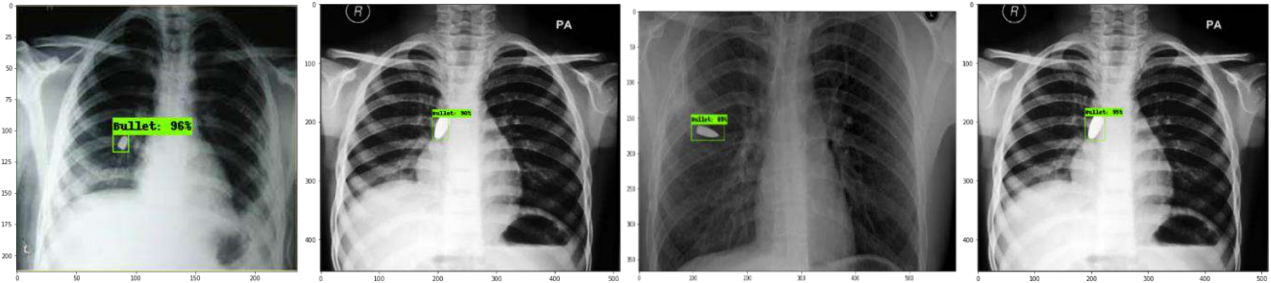

*Figure 7. Chest Xray Predictions*

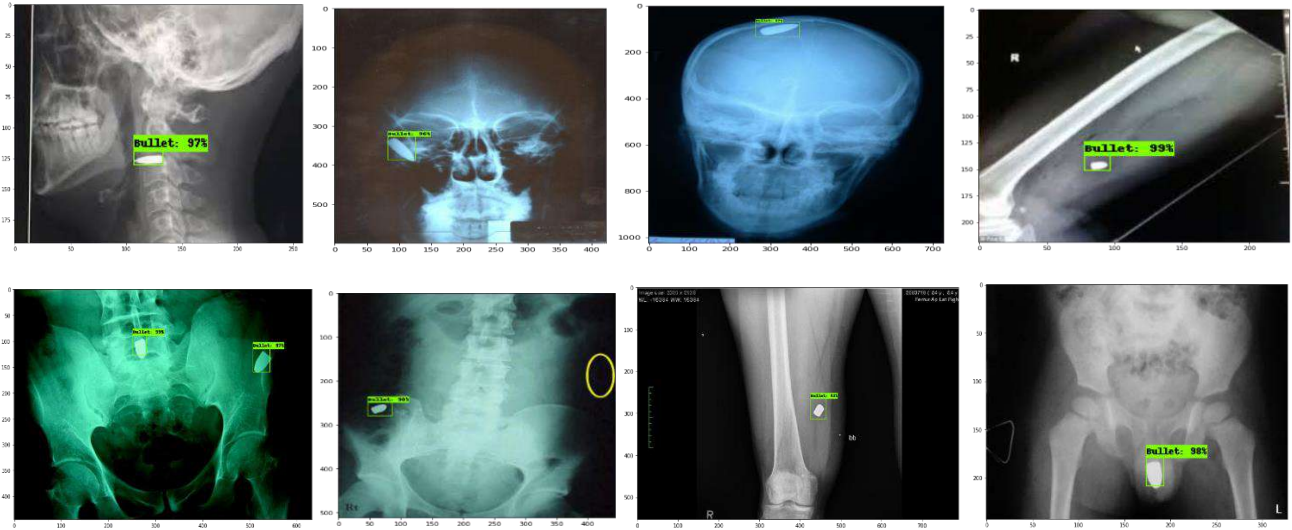

*Figure 8. Other Organs Xray Predictions*

The mean average precision of Faster RCNN in Figure 5 is better than both SSD models and one of the reasons can be that we already discussed that Faster RCNN is better to detect small objects than SSD. Faster RCNN catches smaller objects because of its region proposal network configuration. The mean average precision of each model that showed in graphs is of the last or final execution of the model. Similarly, the average recall of Faster RCNN Resnet 101 COCO in Figure 6 outclass other SSD models' performance, presented in Table 1.

| Model | mAP | AR |
|---|---|---|
| Faster RCNN Resnet101 Coco | 22 | 18 |
| SSD Mobile Net V1 Coco | 9 | 8 |
| SSD Mobile Net V2 Coco | 8 | 8 |

Table 1: *mAP and AR of models (train = 30k, eval= 10k)*

As a comparative study, multiple object detection models have been used to compare each model's performance regarding multiple evaluation metrics (Table 1). These models pre-trained on COCO (common objects in context) dataset (Lin et al., 2014). COCO is designed for multiple aspects such as object detection, segmentation, and captioning. It has 200k plus labeled images. Along with labeled images, it also having 80 object categories.

For each model, test cases were performed not only on chest cases but also on other GSW radiographs such as head, neck, leg, and abdomen presented in figure 7 and 8. It must be noted that all above given results are based on actual radiographs. Improvements can be done by tuning hyperparameters and increasing the number of steps per model. The IOU (intersection over union) value in each model was limited to 50. If the threshold value of the predicted bounding box exceeds the IOU limit of 50 then the model considered it as accurate and generated a bounding box.

The parameters of each model can be changed in the pipeline config file. The label of the class is stored in the labelmap.txt file. As it is difficult for object detection models to detect small objects, so the mean average precision value manifested by the object detection model is low.

## 5. Conclusion

In the proposed study, we have worked on a novel aspect of medical imaging by using deep learning. For the analysis of chest radiographs affected by gunshot wounds, the proposed model gives a good accuracy by using the FPT technique for VGG16 classification. The proposed study used SSD and Faster RCNN object detection models to detect bullets in x-ray scans for localization of bullets. A comparative study was conducted on different object localization models to check each model's accuracy. This work might assist the radiologists in the classification and localization of chest X-rays with fewer human interventions that can save time and cost. In addition to this, it also enlightens researchers to work on this study which was not conducted earlier due to some reasons. The limitation in the study is the lack of the dataset, if this will be incorporated in the future, then much better results are expected.

For future perspectives, the proposed study can be extended by using the other object localization models to improve the comparative study analysis and elaboration of models which will outclass others. The other thing that can also be ensured is the usage of triaging technique and determining which patient needs urgent treatment. This improvement will be allowed based on the number of bullets detected inside the patient's body regarding all other patients. The more the number of bullets inside the sufferer's body, the more urgent the need to cure the body.